# Classification Of Fake News Headline Based On Neural Networks

Ke Yahan[1],
Northwestern Polytechnical University,
Xi'an, China,
keyahan@mail.nwpu.edu.cn,

Ruyi Qu[1]
University of Toronto
Toronto,Canada
ruyi.qu@mail.utoronto.ca

Lu Xiaoxia[2]
Sun Yat-sen University,
Guangzhou, China
luxiaoxia2018@outlook.com

**Abstract—** Over the last few years, Text classification is one of the fundamental tasks in natural language processing (NLP) in which the objective is to categorize text documents into one of the predefined classes. The news is full of our life. Therefore, news headlines classification is a crucial task to connect users with the right news. The news headline classification is a kind of text classification, which can be generally divided into three mainly parts: feature extraction, classifier selection, and evaluations. In this article, we use the dataset, containing news over a period of eighteen years provided by Kaggle platform to classify news headlines. We choose TF-IDF to extract features and neural network as the classifier, while the evaluation metrics is accuracy. From the experiment result, it is obvious that our NN model has the best performance among these models in the metrics of accuracy. The higher the accuracy is, the better performance the model will gain. Our NN model owns the accuracy 0.8622, which is highest accuracy among these four models. And it is 0.0134, 0.033, 0.080 higher than its of other modes.

## I. Introduction

Over the last few years, text classification has been widely applied in many fields, due to the developing of machine learning [1, 2]. There are 85k news articles to users in Europe alone, which means our life is full of textual data in many languages and contexts.[3] Therefore, news headlines classification is a crucial task to connect users with the right news. The news headline classification is a kind of text classification, which can be generally divided into three mainly parts: feature extraction, classifier selection, and evaluations.

In this article, we will use the dataset, containing news over a period of eighteen years provided by Kaggle platform to classify news headlines. We choose TF-IDF to extract features and neural network as the classifier, while the evaluation metrics is accuracy.

The rest of article contains four parts: section II presents related work about text classification. Section III provides the methodology we adopt, while in section IV the experiments result and compared experiments are presented. Finally, the section V concludes the whole work of this article.

## II. Introduction

Traditionally, for text classification, bag-of-words model [3] was used to represent a document. This simple approach uses the term frequency as features followed by a classifier such as Naïve Bayes [4] or Support Vector Machine (SVM) [5]. One drawback of this approach is that it ignores the word order and grammatical structure. It also suffers from data sparsity problem when the training set's size is small but has shown to give good results when size is not an issue.

Therefore, feature selection is typically carried out to reduce the effective vocabulary size by removing the noisy features [6] before these methods. One key property of these linear classifiers is that they assign high weights to some class label specific keywords, which are also known as lexicons.

The next generation of approaches includes neural networks that have shown to outperform bag-of-words models in text classification tasks [7] These methods typically use multiple layers of Convolutional Neural Network (CNN) [8] and/or Long Short-Term Memory (LSTM) networks [9]. The motivation of using these complex neural network approaches for classification tasks comes from the principle of compositionality [10], which states that the meaning of a longer expression (e.g., a sentence or a document) depends on the meaning of its constituents. It is believed that lower layers of the network learn representations of words or phrases, and as we move up the layers more complex expressions are represented.

- *Our Contribution*
- A neural network model for classification is proposed by us.
- We choose TF-IDF as feature extraction.
- We introduce experimental results and apply accuracy as metrics, comparing our model with other models on the same dataset provided by Kaggle.

### III. METHODOLOGY

> TF-IDF

Term Frequency-Inverse Document Frequency [11] (TF-IDF) is a common weighting technique for information retrieval and information exploration. TF-IDF is a statistical method used to assess the importance of a word to a file set or to one of the documents in a corpus.

TF (Term Frequency) indicates how often a term appears in text, a number that is usually normalized (usually the frequency of words divided by the total number of words in an article) to prevent it from favoring long files (the same word may have a higher frequency of words in a long file than in a short file, regardless of whether the word is important or not). TF is represented by a formula as follows:

$$TF_{i,j} = \frac{n_{i,j}}{\sum_k n_{k,j}}$$

Where $n_{i,j}$ represents the number of times a term $t_i$ appears in document $d_j$, $TF_{i,j}$ is the frequency at which term $t_i$ appears in a document $d_j$.

IDF (Inverse Document Frequency) indicates the prevalence of keywords. If you have fewer documents that contain the entry $i$, the larger the IDF, which means that the term has good category differentiation. The IDF formula is as follows:

$$IDF_i = \log \frac{|D|}{1 + |j: t_i \in d_j|}$$

Where $|D|$ represents the number of documents, $|j: t_i \in d_j|$ is the number of documents that contain the entry. To prevent the occurrence of an operation error by including a number of terms $t_i$ is 0, it plus 1.

> Classifier

Deep neural networks (DNN) are designed to learn by multi-connection of layers that every single layer only receives the connection from previous and provides connections only to the next layer in a hidden part. The input layer is constructed via TF-IDF. The output layer is equal to the number of classes for binary classification. In our task, what we do is a binary classification, means we just classify the fake news and real news. The structure our Neural networks are shown in the following figure 2.

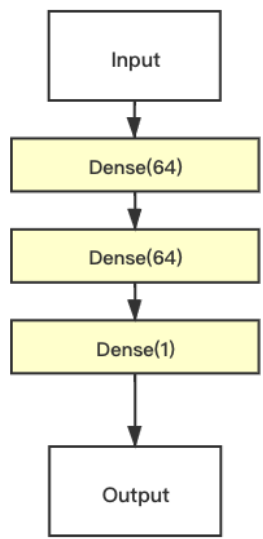

figure 2: the structure of neural network

### IV. Experiment

In this section, we will discuss about the result and compared experiments.

- **Experimental Data**

The dataset is a combination of 3 singular datasets: A Million News Headlines, Fake and real news, Getting Real about Fake News, where the A Million News Headlines dataset will be labeled as real news headlines, and the other two will be labeled as fake news. There are two columns in Million News Headline, which are date and headline. while Fake and real news dataset contains four description columns: title, text, subject and date. Getting Real about Fake News has eight columns, and they are orders, author, date, title, text, language… We extract the same feature columns between these three datasets, contact them together, and get 50000 real headlines, 29601 fake headlines. The word cloud is shown in figure 1.

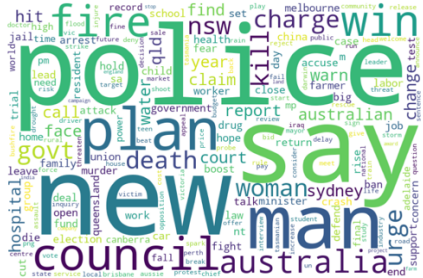

Figure 1: word cloud

- **Data preprocessing**

We use nltk and genism to help us do the data preprocessing. The preprocessing can be divided into Tokenization, Stop Words removing, Stemming. Tokenization [] breaks a stream of text into words, phrases, symbols, or other meaningful elements called tokens. Text and document classification include many words which do not contain important significance to be used in classification algorithms called stop words []. In addition, one word could appear in different forms (i.e., singular and plural noun form) while the semantic meaning of each form is the same. One method for consolidating different forms of a word into the same feature space is stemming [].

- **Experimental settings**

| loss | 'binary_crossentropy' |
| --- | --- |
| optimizer | adam |
| epochs | 50 |
| batch_size | 512 |
| Learning rate | 1e-4 |

- **Metrics**

In our task, the important metrics we focus is the accuracy, which reflects the difference between our prediction and the ground truth. The definition of the accuracy is as follows:

$$accuracy = \frac{(TP + TN)}{(TP + FP + TN + FN)}$$

To know the definition of accuracy we need to grasp the definitions of TP, FP, TN and FN. The confusion matrix is shown in figure 3.

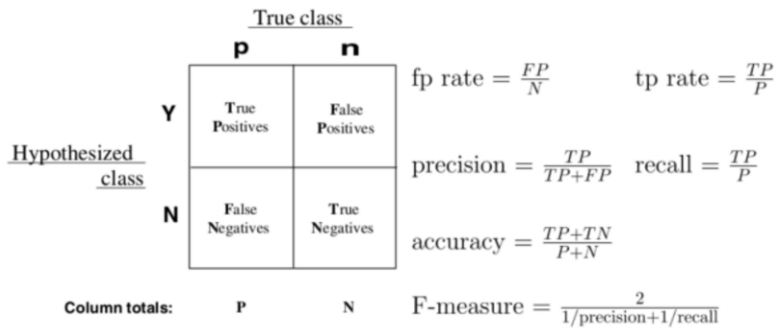

Figure 3: the confusion matrix

- **Experimental Results and Analyses**

We do the compared experiments with other classical models, such as decision tree, random forest, support vector classification using the same dataset and metrics.

| Models | ACC |
| --- | --- |
| **Decision Tree** | 0.8488 |
| **Random Forest** | 0.8589 |
| **SVC** | 0.8542 |
| **NN** | 0.8622 |

It is obvious that our NN model has the best performance among these models in the metrics of accuracy. The higher the accuracy is, the better performance the model will gain. Our NN model owns the accuracy 0.8622, which is highest accuracy among these four models. And it is 0.0134, 0.033, 0.080 higher than its of other modes.

V. Conclusion

Over the last few years, Text classification is one of the fundamental tasks in natural language processing (NLP) in which the objective is to categorize text documents into one of the predefined classes.

In this article, we use the dataset, containing news over a period of eighteen years provided by Kaggle platform to classify news headlines. We choose TF-IDF to extract features and neural network as the classifier, while the evaluation metrics is accuracy. From the experiment result, it is obvious that our NN model has the best performance among these models in the metrics of accuracy. The higher the accuracy is, the better performance the model will gain. Our NN model owns the accuracy 0.8622, which is highest accuracy among these four models. And it is 0.0134, 0.033, 0.080 higher than its of other modes.